\documentclass[a4paper, 12pt, leqno]{article}
\usepackage{enumitem}
\usepackage{url}
\usepackage[pagewise]{lineno} 
\usepackage{amsmath,amssymb}
\usepackage[round]{natbib}
\usepackage{pifont}
\bibpunct{(}{)}{,}{a}{}{,}
\usepackage[utf8]{inputenc}
\usepackage{mathtools}
\usepackage{amsthm}
\usepackage{mathrsfs}
\usepackage{setspace}
\usepackage{dirtytalk}

\renewcommand{\say}[1]{\lq#1\rq}

\usepackage{hyperref}
\begin{document}

\title{What are the odds? Risk and uncertainty about AI existential risk}
\author{Marco Grossi}
\date{January 23, 2025}
\maketitle
\begin{abstract}
This work is a commentary of the article \href{https://doi.org/10.18716/ojs/phai/2025.2801}{AI Survival Stories: a Taxonomic Analysis of AI Existential Risk} by Cappelen, Goldstein, and Hawthorne. It is not just a commentary though, but a useful reminder of the philosophical limitations of \say{linear} models of risk. The article will focus on the model employed by the authors: first, I discuss some differences between standard Swiss Cheese models and this one. I then argue that in a situation of epistemic indifference the probability of P(D) is higher than what one might first suggest, given the structural relationships between layers. I then distinguish between risk and uncertainty, and argue that any estimation of P(D) is structurally affected by two kinds of uncertainty: option uncertainty and state-space uncertainty. Incorporating these dimensions of uncertainty into our qualitative discussion on AI existential risk can provide a better understanding of the likeliness of P(D).
\end{abstract}

\section{Structural relations between layers}
\subsection{The model}
\cite{cappelenAIsurv} analysis makes use of a Swiss cheese model to estimate P(D). The model is based on the following \say{layers} of protection:
\begin{enumerate}
    \item Technical Plateau: Scientific barriers prevent AI systems from becoming extremely powerful.
    \item Cultural Plateau: Humanity bans research into AI systems becoming extremely powerful.
    \item Alignment: Extremely powerful AI systems do not destroy humanity, because their goals prevent them from doing so.
    \item Oversight: Extremely powerful AI systems do not destroy humanity, because we can reliably detect and disable systems that have the goal of doing so.
\end{enumerate}
The authors comment on each possibility in detail, analysing the scope and limitations of each layer, with very rich and useful references.
\par
Swiss-Cheese models originate from \cite{reason1997managing}. The model as intended by Reason should be read as follows:
\begin{itemize}
    \item Slice = protection barrier
    \item Hole = failure
    \item Arrow = path leading to accident. 
\end{itemize}
For an accident to occur, the arrow must pass through each hole. The authors argue that the strength of a Swiss-Cheese model is based on two factors: \say{how reliable each layer of safety is, and on whether the \say{holes} in each layer of safety are independent of the others}. Both factors are thoroughly discussed in the subsequent sections, where the authors argue that \say{each survival story faces its own challenges, which are structurally independent of challenges to other stories} \citeyearpar[3]{cappelenAIsurv}.
\par
Two other factors that are not mentioned are the number of layers and the independence of the layers themselves. The more the layers, the less probable an incident is, assuming that the holes in each layer are independent. Also, if two layers are not independent of each other, then it could be that one fails to exist if another fails to exist. So, while the model assesses the risk by assuming that four layers are in place, there might only be two. 
\par
Consider Covid as an example. Possible layers of security to stop Covid were:
\begin{enumerate}
    \item Border closure.
    \item Isolation for arrivals.
    \item Physical distancing inside the house if one member is sick.
    \item Limits of movement between members of the household (e.g., only one member goes shopping once a week for groceries).
    \item Handwashing
    \item Use of masks.
\end{enumerate}
Layers 1 and 2 were enacted by the government, 3 and 4 by households, 5 and 6 by individuals. Part of the robustness of the model lies in the fact that these three blocks work independently: the fact that border closure is or is not in place doesn't affect whether I was my hands frequently, or whether we decide to go shopping only once a week, and vice-versa.
\par
The same situation doesn't hold in the model in \citet{cappelenAIsurv}, where different layers are defined so that the activate only if the previous ones fail. The authors are very clear on this point. When they discuss Oversight they note the following: 
\begin{quote}
    We will make the simplifying assumption that oversight is incompatible with the other stories. [..] this definitional choice is not substantive. When we estimate the overall probability of survival, we will think of our four stories as being four layers of safety. We will consider the chance that a given layer succeeds on the supposition that previous layers fail. So in practice, we may as well define the survival story as implying that the previous layers fail. \citeyearpar[15]{cappelenAIsurv}
\end{quote}
Call Technical Plateau \say{T}, Cultural Plateau \say{C}, \say{Alignment} A, and Oversight \say{O}. The author's point that this definitional choice is \say{not substantive} can be based on the observation that the probability of two events $A$ and $B$ happening is $P(A)\times P(B/A)$. So, if Doom happens if all layers fail, then what we aim to calculate is the probability of $\neg T \& \neg C \& \neg A \& \neg O$, which by iterated applications of the rule just stated must be the following:
\[ \tag{D1}P(D) = P(\neg T)\times P(\frac{\neg C}{\neg T}) \times P(\frac{\neg A}{\neg T \& \neg C}) \times P(\frac{\neg O}{\neg T \& \neg C \& \neg A})\]
This formula is exactly the one used by the authors. 
\subsection{Evaluating P(D) under epistemic indifference}
While it doesn't change the equation for P(D), I claim that whether the layers are interdependent can matter when we try to estimate the probability of P(D). For example, consider Alignment and Oversight in general, and set aside the definitional choice that makes Oversight incompatible with Alignment. If AI becomes extremely powerful, it is quite likely that the methods we use to control it will make use of AI itself, as AI will become pervasive. In particular, it is likely that the best shot we have at controlling AI is with AI itself. Thus, if AI can be made aligned, then AI itself would be a reliable method for Oversight, because AI would oversee itself.\footnote{For as similar point see \citealt{SalibAIwillnot}} Let's distinguish between Oversight methods that presuppose the use of AI, and thus presuppose that AI is aligned in some way, and those that do not. Call the first $O_1$ and the second $O_2$. $P(O_1/A)= 1$, thus $P(O_1/\neg A)$= 0.
\par
Now suppose we are epistemically indifferent toward each layer of the model. Without any further information, we might deem rational to attribute to each layer a 50\% chance of success, conditional on the previous layer failing. As the authors note in section 5, this gives the following results for P(D), by applying D1:
\[P(D) = 0.5 \times 0.5 \times 0.5 \times 0.5 = 6.25\%\]
Yet we have failed to recognise that Alignment is itself a method of Oversight, and thus Oversight should be broken down into $O_1$ and $O_2$. Given the authors' definitional choice, $O_2$ is actually what they call \say{Oversight}, which is incompatible with Alignment. Yet, it should be clear by now that we cannot give it a 50\% chance of succeeding modulo the failure of Alignment, even if we are in a state of epistemic indifference. We should rather refine our probabilities: $O= O_1\lor O_2$. Since I am indifferent between $O_1$ or $O_2$, $P(O_1/O)=P(O_2/O)=0.5$. $P(O/\neg A)$ is now only $25\%$, since $P(O_1/\neg A)=0$. Thus $P(\neg O/\neg A)=75\%$. Applying D1:
\[0.5 \times 0.5 \times 0.5 \times 0.75 = 9.375\% \]
That's roughly a 33\% increase in the chances of Doom, which is significant.\footnote{My argument relies on the Principle of Indifference which is controversial.} Note that I am still in an epistemic state of indifference, the only variation is that I have refined my probability field. The prospects of Doom would be even bleaker if we further believe that, in case AI becomes sufficiently powerful, it will be pervasive and therefore our best shot at controlling it is through AI itself. For then we would give $P(O_1/O)$ a higher probability than $P(O_2/O)$, and therefore a lower overall probability to $P(O/\neg A)$.
\par
To make my point clearer, I will use the author's own analogy of a boat rescue: \say{imagine that you’re stranded on a desert island. Your survival depends on either being picked up by a boat from company 1 or a boat from company 2. Each company has a chance of sending a boat to your island, and a chance of stranding you}. \citep[8]{cappelenAIsurv}. Call B1 and B2 the event that company 1 sends a boat and company 2 sends a boat, respectively. The probability of not being rescued is $\neg B1 \& \neg B2/\neg B1$. In a situation of total ignorance you would place a 0.25\% probability to it: 0.5 times 0.5. Yet now suppose that you know that if company 1 doesn't send a boat, it is likely because they don't have a boat at all to send in a reasonable vicinity of the island. You also know that there are two possibilities when company 2 sends a boat: either if a boat of company 1 intercepts a boat of company 2 and make them aware of your predicament, or if they intercept your SOS signal. Your probability of survival has now dropped, since $P(\neg B2/\neg B1)$ is higher than 0.5.
\par
A similar point may be raised with respect to Cultural Plateau and Oversight. Again, let set aside the definitional choice that makes the two incompatible and consider the problem of Oversight and Cultural Plateau in general. A crude but effective method to detect/disable AI systems that have the goal of destroying humanity is to successfully ban research into powerful AI systems in general. Thus, if Oversight fails, even without the definitional choice, then we don't have a method to control AI systems that want to destroy humanity, which means that we don't have a crude method either, so Cultural Plateau must have failed. 
\par
We might introduce this refinement in the probability field by redefining $O$ as $O_1 \lor O_2 \lor O_3$, where $O_1$ are methods that involve alignment, $O_2$ are methods that involve Cultural Plateau, and $O_3$ are methods that involve neither of those. Alignment presupposes that AI systems become extremely powerful, otherwise they cannot align: $A$ and $C$ are incompatible. Thus, $O_1$ and $O_2$ are also incompatible, since $O_1$ implies $A$ and $O_2$ implies $C$. Suppose I am epistemically indifferent between $O_1$, $O_2$, or $O_3$. Since $P(O_1/\neg A)=0$ and $P(O_2/\neg C)=0$, the probability of $O/\neg C \& \neg A$ goes from 0.5 before the refinement to $0.5/3$. So $P(\neg O/\neg C\& \neg A)= 0.8(3)$ Applying D1:
\[0.5 \times 0.5 \times 0.5 \times 0.8(3) = 10.416\%\]
In any case, even in a state of epistemic uncertainty the risk of P(D) is significantly higher than 6.25\%.
\section{Risk vs uncertainty}
Uncertainty is ubiquitous in models that attempt to describe complicated social phenomena. By \say{uncertainty} I mean risk that is unquantifiable and simply does not show itself in the model. The definition goes back to \cite{knightrisk} and \cite{keynes1937general}, who distinguished between \say{risk} -- a situation where the odds are known but not 1 or 0 -- and uncertainty -- situation where the odds are unknown, either because we cannot have access to them or because they are undefined. The distinction between risk and uncertainty is philosophically dubious since clearly uncertainty is a kind of risk, but it has become standard in decision theory and Economics \citep{luce2012games}. Uncertainty, being a form of risk, is relevant in the assessment of the risk of P(D), yet it doesn't show itself in the model. 
\par 
Swiss Cheese models are linear: each barrier acts on its own, the holes are independent, and the path to an accident is also linear, passing through each layer in a specific sequential order. Their graphic simplicity is part of their allure, but it lacks nuance. The system we aim to represent is influenced by social, cultural, technological, and political factors, and AI is a rapidly changing digital technology, and thus a constantly moving target. Therefore, the  \say{holes in the cheese} themselves are constantly moving as time goes by \cite{shorrockwhomoved}. If we identify \say{failure} with Doom, it is quite likely that any \say{failure} in such a complex system will be caused by systemic factors which are non-linearly related \citep{levenson1995soft}. Accidents in complex systems often happen because different sub-systems interact with each other in unexpected ways. Thus \say{safety} is a systemic property, not a property of single layers \citep{leveson2011applying}. 
\par
Given the intrinsic limitations of Swiss Cheese models, there are types of risk that are simply not considered, but are relevant to P(D). I identify two main types of uncertainty in this model, borrowing the terms from \cite{bradley2014types}:
\begin{itemize}
    \item Option Uncertainty (OU)
    \item State-Space Uncertainty (SU)
\end{itemize}
\subsection{State-Space Uncertainty}
Let's start with the second type, which is discussed by the authors at the end of the paper. Since we are dealing with a complex social system, we probably don't know what the relevant possibilities about P(D) are, and whether the taxonomy provided by the authors is exhaustive. To address this issue, the authors briefly suggest adding another option for \say{other survival stories}. This is often referred to as \say{residuum} or \say{catch-all hypothesis} \citep{wenmackers2016new, hansson2022can}. CH is the best known way of incorporating SU in the model, but it faces different challenges. I will talk about two of them: the No-calibration problem and the Unknown Relations problem. 
\par
Let's start with the No calibration problem. Since by definition, CH is made of unknown unknowns, we have no idea how many other survival stories are out there, so there is no way of calibrating CH. What these stories will look like and which layers of protection we will need to avoid AI Doom will depend on what AI capabilities, but forecasting the uses and capabilities of new technologies is often impossible. The steam engine was developed in the eighteenth century primarily as a tool for removing water from flooded mines; Bell Laboratories initially denied the application for a patent of the laser, as they saw no possible application of it to the telephone industry \citep{rosenberg1995technology}. History is full of examples where there was simply no way of forecasting how a technology would develop.
\par
Now suppose we are indifferent towards CH, so we decide to give it a 50\% probability. Modulo a 50\% chance to each layer failing conditional on the previous ones failing, following D1, P(D) gets halved from 6.25\% to 3.125\%. Yet, suppose we then discover that under CH there are actually four equally possible alternative survival stories. From a condition of ignorance, we should give them 50\% probability to each failing, so now P(D) get shrunk to 0.39\%. The moral of the story is that the variation in the possible values of CH is so large that there is no sensible basis for giving it neither a specific number nor range \citep{shimony1970}. Since P(D) depends on P(CH), it is likely that we cannot  give neither a specific number nor range to P(D) either, but only a conditionalised number based on CH \citep{wenmackers2016new}. 
\par
Let's discuss now the Unknown Relations problem. A related issue with CH is that we have no idea how components of different unknown stories might interact with each other. It might be that a failure of some subset of them influences the probability of failure of other layers. So, even if the authors are correct in claiming that \say{each survival story faces its own challenges, which are structurally independent of challenges to other stories}, this holds true of the stories discussed in the paper, not of the unknown stories in CH. It follows that any probability will likely need to be conditionalised to CH, not just the probability of Doom.
\subsection{Option Uncertainty}
Option Uncertainty is uncertainty about what consequences a certain option will have. In the model in \citet[3]{cappelenAIsurv}, for each layer we have two options: \say{Yes} and \say{No}. \say{Yes} leads to \say{Survival}, \say{No} leads to \say{Destruction}. Of course this is a simplification, and a useful one, but we have to remind ourselves that reality is not Model Land. It is likely that any survival or doom story will be non-linear and without any specific root cause. Reflexivity and feed-back loops further complicate the picture. By \say{reflexivity} I mean phenomena where our assessment of risk influences the risk itself. I will give two examples where this might happen in the model. In particular, us attempting Cultural Plateau will affect our chances of Oversight and our chances of Alignment.
\par
\citet[11]{cappelenAIsurv} cite the possibility of warning shots as a possible trigger for Cultural Plateau, and give the example of nuclear energy: \say{Nuclear disasters like Chornobyl, Fukushima, and Three Mile Island had a significant impact on the ability to build nuclear reactors}. However, one might argue that, precisely because research in nuclear energy was halted and funds dried out after these disasters, we didn't develop fast enough alternative ways of building nuclear reactors which would have been much safer, and today we are still using \say{old technology} which is less safe and that we can control less. Therefore, us assessing the risk of nuclear disaster led to a decision that increased the risk of nuclear disaster. 
\par 
A similar argument could be made for AI existential risk. Consider the following scenario:
\begin{quote}
A warning shot event comes about. Our increased awareness of the risk that AI poses pushes humanity towards Cultural Plateau. Funds for AI development and research dry up. However, in the end, a powerful AI gets created anyway, albeit more slowly and mostly through the recursive self-improvement of AI itself.
\end{quote}
Similarly to the nuclear case, in this scenario AI becomes extremely powerful mostly through recursive self-improvement, rather than through our understanding of it. Since we haven't \say{kept up} with AI research because we tried to implement a Cultural Plateau, the probability of Oversight is lowered, as our methods to control AI are likely more outdated. Thus the risk of Doom is higher. This situation is not really represented in the model: the point is not that Cultural Plateau fails, but that us trying to reach Cultural Plateau impairs our ability to oversee AI later, in case Cultural Plateau fails. This complicates the issue of finding an optimal safety strategy.
\par
A similar point can be made for Cultural Plateau and Alignment. I will use a story inspired to the \say{Roko's basilisk} thought experiment. In the original, gory version, a powerful AI god starts torturing anyone who imagined its possible coming but didn't actively help bring it about \citep{rokobasilisk}. We can draw a similar story here. Consider again the scenario above. This powerful AI agent that eventually gets created would have an incentive to dis-align with humanity, since by analysing past data it can see humanity actively tried to prevent its existence, and thus will likely conclude that humanity is a threat to the AI's own survival. Again, us trying Cultural Plateau and then fail impacts the probability of Alignment.
\section{Conclusion}
To approach a problem we have to start from somewhere. \cite{cappelenAIsurv} provide a useful model and a useful structure to start thinking about the probability of AI Doom in a systematic way. More work needs to be done to try to incorporate uncertainty into the model. This will be especially relevant for deciding which is the optimal strategy for survival.

\bibliographystyle{apalike}
\bibliography{bibl}
\end{document}